\documentclass[sigconf,natbib=true]{acmart}%%

\usepackage{amsthm}
\usepackage{amsmath}
\usepackage{algorithm}
\usepackage{enumitem}
\usepackage{picinpar}
\usepackage{lineno}
\usepackage{graphicx}

\usepackage{algorithmicx}
\usepackage[noend]{algpseudocode}
\usepackage{subfigure}
\usepackage{multirow}
\usepackage{color}
\usepackage{balance}
\usepackage{enumitem}
\usepackage{hhline}
\usepackage[normalem]{ulem}
\usepackage{booktabs}
\usepackage{wrapfig}
\usepackage{cancel}
\usepackage{hyperref}
\usepackage{makecell}

\newcommand{\bL}{\ensuremath{\mathcal{L}}}

\newcommand{\bN}{\ensuremath{\mathcal{N}}}

\renewcommand{\vec}[1]{\ensuremath{\mathbf{#1}}}

\newcommand{\stitle}[1]{\vspace{1mm} \noindent {\bf #1}}

\newcommand{\method}[1]{\textsc{#1}}
\newcommand{\model}{\method{EVP}{}}

\newcommand{\eat}[1]{}

\newcommand{\stkout}[1]{\ifmmode\text{\sout{\ensuremath{#1}}}\else\sout{#1}\fi}

\AtBeginDocument{%
  }

\setcopyright{acmlicensed}
\copyrightyear{2026}
\acmYear{2026}
\acmDOI{XXXXXXX.XXXXXXX}
\acmConference[Conference acronym 'XX]{Make sure to enter the correct
  conference title from your rights confirmation email}{June 03--05,
  2026}{Woodstock, NY}
\acmISBN{978-1-4503-XXXX-X/2018/06}
\usepackage{multirow}
\begin{document}

\title{Event-Aware Prompt Learning for Dynamic Graphs}

\renewcommand{\shortauthors}{Trovato et al.}

\author{Xingtong Yu}
\affiliation{%
 \institution{The Chinese University of Hong Kong}
  \city{Hong Kong}
  \country{China}}
\email{xtyu@se.cuhk.edu.hk}

\author{Ruijuan Liang}
\affiliation{%
 \institution{University of Science and Technology of China}
 \city{Hefei}
 \country{China}}
\email{lrjuan@mail.ustc.edu.cn}

\author{Renhe Jiang}
\affiliation{%
 \institution{The University of Tokyo}
 \city{Tokyo}
 \country{Japan}}
\email{jiangrh@csis.u-tokyo.ac.jp}

\author{Dongyuan Li}
\affiliation{%
 \institution{The University of Tokyo}
 \city{Tokyo}
 \country{Japan}}
\email{lidy94805@gmail.com}

\author{Yunxiao Zhao}
\affiliation{%
 \institution{Shanxi University}
 \city{Shanxi}
 \country{China}}
\email{yunxiaomr@163.com}

\author{Xinming Zhang}
\affiliation{%
  \institution{University of Science and Technology of China}
  \city{Hefei}
  \country{China}}
\email{xinming@ustc.edu.cn}

\author{Yuan Fang}
\affiliation{%
  \institution{Singapore Management University}
  \city{Singapore}
  \country{Singapore}}
\email{yfang@smu.edu.sg}

\begin{abstract}
  Real-world graph typically evolve through a series of events, modeling dynamic interactions between objects across various information retrieval applications. While dynamic graph neural networks have emerged as a popular solution to modeling dynamic graphs, more recent prompt learning approaches offer a parameter-efficient alternative. However, existing approaches mainly operate at the node--time level and fail to explicitly exploit the structural evolutions induced by historical events. 
  In this paper, we propose \model, an event-aware dynamic graph prompt learning framework that can serve as a plug-in to existing methods, enhancing their ability to leverage historical events. First, we extract node-specific event histories, then perform event adaptation via lightweight, task-aligned prompts to modify fine-grained event evidence toward the downstream objective. Second, we propose an event aggregation mechanism to integrate adapted events across time using a recency-aware time-decay prior and a pattern-aware dynamic prompt to capture both short-term dynamics and informative long-range patterns. Extensive experiments on four public datasets demonstrate that \model\ consistently improves over state-of-the-art baselines. Codes are available at \textcolor{blue}{\url{https://anonymous.4open.science/r/EVP-F57E/}} for anonymous reviewing.
\end{abstract}






\maketitle

\section{Introduction}\label{sec.intro}
Dynamic graphs capture evolving graph structures driven by different temporal events, and underpin diverse information retrieval applications. Examples of such temporal events include a user posting new blogs on Reddit \cite{kumar2018community,iba2010analyzing}, which drives timely content recommendation in social media; creating new pages on Wikipedia \cite{kumar2015vews}, which enables the search and retrieval of emerging topics; or listening to new music genres \cite{yu2024dygprompt}, which facilitates personalization under shifting user interest. 

%Graphs are widely used to model interactions among different entities in various applications \cite{xia2021graph,wang2021graph}. Real-world graph structures generally evolve over time driven by different temporal events. For example, a user posts a new blog on Reddit \cite{kumar2018community,iba2010analyzing}, creates a new page on Wikipedia \cite{kumar2015vews}, or listens to a new genre of music \cite{yu2024dygprompt}. Such event streams underpin a wide range of retrieval- and recommendation-centric objectives, including next-interaction prediction, user-state modeling, and personalization under shifting user interests and platform dynamics. 

\begin{figure}[t]
\centering
\includegraphics[width=1\linewidth]{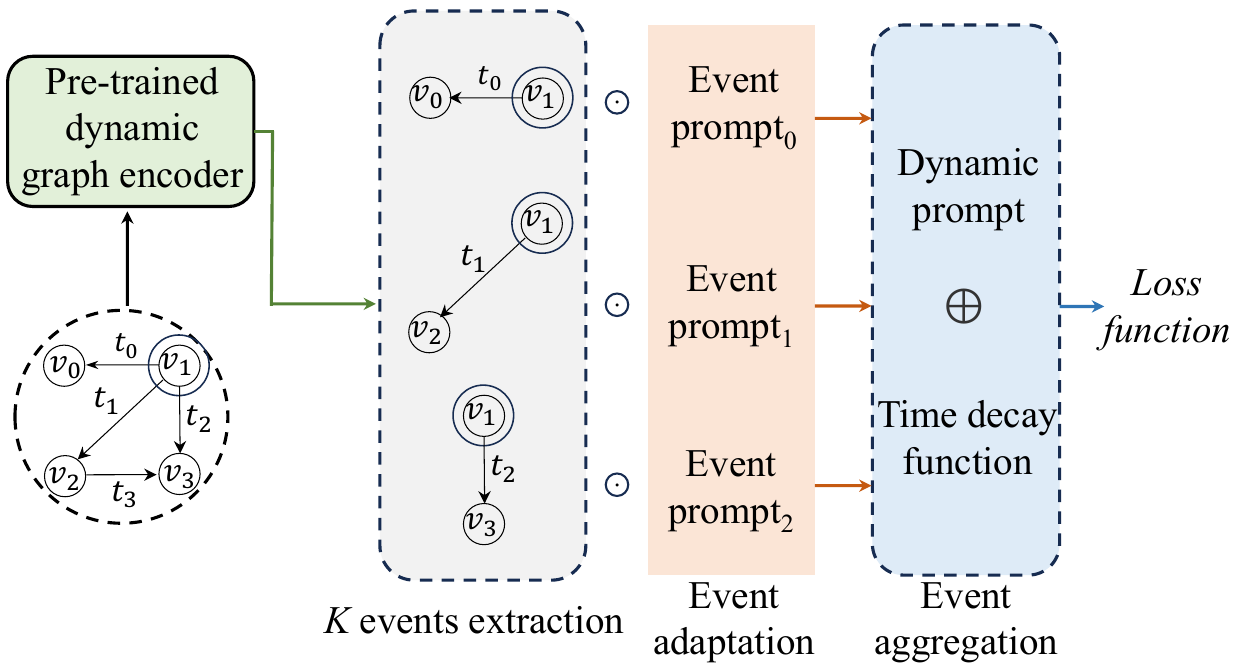}
\caption{Illustration of \model.}
\Description{illustration of the proposed model.}
\label{fig.intro-motivation}
\end{figure}

A mainstream approach for modeling such evolving graphs is dynamic graph neural networks (DGNNs) \cite{rossi2020temporal,xu2020inductive,dubey2025mintt}. These models typically update a node's representation by iteratively aggregating temporal messages from its neighboring nodes. While DGNNs are typically trained on link prediction tasks, the downstream task may differ (such as node classification), leading to a significant gap between pre-training and downstream task objectives. More recent pre-training and fine-tuning strategies on dynamic graphs \cite{bei2024cpdg,chen2022pre,tian2021self} also suffer from a similar limitation. These methods generally pre-train a model using self-supervised signals derived from intrinsic graph properties, and then fine-tune the model on task-specific labels in downstream tasks. Nevertheless, fine-tuning can be costly, and the learned representations may still be insufficiently aligned with downstream decision boundaries, especially when the supervision target shifts away from the pre-training objective. Moreover, both DGNN and pre-training paradigms tend to absorb historical events only implicitly through message passing or pretext tasks, leaving explict historical event knowledge under-exploited during downstream adaptation.

To bridge the objective gap in a parameter-efficient manner \cite{liu2023pre}, prompt learning has been applied to static graphs \cite{liu2023graphprompt,sun2022gppt,sun2023all}. They modify the node features or embeddings by introducing lightweight, task-specific prompts, which are then tuned for downstream tasks. This approach is particularly efficient in low-resource scenarios as only the prompt parameters are adjusted, while the pre-trained encoder remains frozen. However, these static prompt-based methods are unable to capture the temporal dynamics inherent in evolving graphs. Recently, prompt learning has been extended to dynamic graphs \cite{yu2024dygprompt,chen2024prompt}, where time-aware and node-aware prompts are used to model interactions between nodes and time. Despite this progress, existing dynamic prompt methods only capture temporal dynamics at the node--time level and fail to explicitly leverage the structural evolutions induced by different historical events during adaptation. In many systems, however, events are the atomic drivers of graph evolution and provide the most direct behavioral evidence; overlooking event-level knowledge can therefore limit adaptation effectiveness.
To solve these limitations, in this work, we propose an \textbf{EV}ent-aware dynamic graph \textbf{P}rompt learning method, \model, as shown in Fig.~\ref{fig.intro-motivation}. \model\ leverages historical events as first-class adaptation signals and serves as a plug-in that can enhance existing dynamic graph learning pipelines, including traditional DGNNs, dynamic graph pre-training methods, and dynamic graph prompt learning methods. However, the realization of \model\ is non-trivial due to two key challenges.

First, \textit{how can we adapt fine-grained event evidence to heterogeneous downstream objectives without fine-tuning the backbone?}
Dynamic graphs evolve through timestamped events, yet pre-trained encoders typically learn event signals under link-oriented objectives, while downstream tasks may require different decision boundaries and attend to different aspects of interaction evidence. Although previous prompt learning work on graphs \cite{liu2023graphprompt,sun2022gppt,sun2023all} bridges the task objective gap and dynamic prompt learning captures interactions between nodes and time \cite{yu2024dygprompt,chen2024prompt}, they struggle to explicitly leverage the structural evolutions driven by different historical events. To address this gap, \model\ introduces an \emph{event adaptation} mechanism that modifies each event embedding via lightweight, task-aligned prompts. This transforms raw historical events into downstream-relevant representations, improving alignment between pre-trained knowledge and downstream objectives while keeping the encoder frozen.

Second, \textit{how do we aggregate the historical knowledge of events across different time to capture both temporal recency and accumulated patterns?} 
Historical events contribute unevenly to a node's current behavior: recent events often carry immediate relevance (temporal recency), whereas earlier events may expose recurring patterns that reflect long-term preferences (accumulated patterns). For instance, if a user regularly posts about a specific topic, even if that event occurred several days ago, it could be more indicative of future behavior than a recent event related to a different topic. Thus, it is crucial to weigh the importance of individual events and aggregate them into a cohesive representation of the user’s behavior over time. In \model, we introduce an an \emph{event aggregation} mechanism that combines a recency-aware time-decay function with a pattern-aware dynamic prompt to integrate historical events, yielding a compact history-aware representation that captures both short-term dynamics and informative long-range patterns. This allows the model to integrate comprehensive and relevant historical event knowledge across different times, providing a more holistic view of user behavior.

Both the event adaptation and event aggregation mechanisms in \model\ can be seamlessly integrated with existing dynamic graph learning methods. For DGNNs, we can directly employ them as the backbone of \model. For pre-training methods, we utilize the output embeddings from the pre-trained graph encoder to compute event embeddings, and then perform event adaptation and aggregation to incorporate historical event knowledge. For prompt learning methods, node features are modified prior to being input into the pre-trained graph encoder, resulting in prompt-adjusted node embeddings. We then apply our event adaptation and aggregation in the same manner as in the pre-training methods, ensuring consistency across different learning paradigms.

In summary, the contributions of this work are threefold. 
(1) We propose \model, an event-aware dynamic graph prompt learning framework, which could serve as a plug-in to enhance present dynamic graph learning methods with historical event knowledge for downstream tasks.
(2) In \model, we design an event adaptation mechanism to capture the fine-grained characteristics of historical events, and an event aggregation mechanism to integrate comprehensive and relevant historical knowledge for downstream tasks.
(3) We conduct extensive experiments on four benchmark datasets, demonstrating the superior performance of \model\ compared to state-of-the-art approaches.

\section{Related Work}

\stitle{Dynamic graph learning.}
In real-world applications, graph structures generally evolve continuously, necessitating dynamic graph modeling approaches. A mainstream technique for dynamic graph learning is dynamic graph neural networks (DGNNs), which update node embeddings by aggregating time-dependent information from neighboring nodes \cite{skarding2021foundations}. Existing DGNNs adopt different strategies to capture temporal dynamics, including dynamic random-walk based modeling \cite{nguyen2018continuous,wang2021inductive}, temporal encoders coupled with message passing \cite{xu2020inductive,cong2023we,rossi2020temporal,yu2023towards}, and temporal point-process based formulations for event-driven evolution \cite{kumar2019predicting,trivedi2019dyrep,wen2022trend}. While effective for link-oriented objectives, transferring DGNN representations to heterogeneous downstream tasks can be challenging when supervision and goals differ.

\stitle{Dynamic graph pre-training.}
Recently, dynamic graph pre-training techniques have been proposed, following a ``pre-training, fine-tuning'' paradigm. These methods first leverage self-supervised learning techniques—such as structural and temporal contrastive learning \cite{bei2024cpdg,tian2021self,li2022mining}, dynamic graph generation \cite{chen2022pre}, and curvature-adjusted Riemannian graph neural networks \cite{sun2022self}—to learn task-agnostic representations in dynamic graphs. They are then adapted to downstream tasks through fine-tuning.
However, for DGNNs and dynamic graph pre-training methods, a significant gap exists between the pre-training and downstream task objectives, hindering the effective transfer of pre-trained knowledge and limiting the performance on downstream tasks.

\stitle{Dynamic graph prompt learning.}
To bridge the gap between pre-training and downstream tasks, prompt learning was first proposed for static graphs \cite{liu2023graphprompt,sun2023all,fang2023universal}, where lightweight prompts modify node features/embeddings to align a (pre-trained) encoder with downstream supervision in a parameter-efficient manner. Since static prompts cannot capture temporal evolution, recent work has extended prompt learning to dynamic graphs \cite{yu2024dygprompt,chen2024prompt} by introducing time-aware and node-aware prompts to model interactions between nodes and time. Nevertheless, these methods mainly emphasize node--time relationships during adaptation and typically overlook explicit \emph{historical event knowledge}, although event sequences drive graph evolution and may carry fine-grained, time-varying signals useful for downstream tasks. Our work complements prior efforts by developing an event-aware prompt learning framework that leverages historical events as a plug-in to enhance diverse dynamic graph learning pipelines.

\section{Preliminaries}\label{sec.pre}
In this section, we present the essential background and define the scope of our work.

\stitle{Dynamic graph.}
Dynamic graph is defined by \( G = (V, E, T) \), where \( V \) and \( E \) are the set of nodes and edges, respectively, and \( T \) is the timeline. Each edge \( (v_i, v_j, t) \in E \) represents an interaction from nodes \( v_i \) to \( v_j \) at time \( t \), also known as an event. Node feature vector \( \vec{x}_{t,v} \in \mathbb{R}^d \) evolves over time, serves as a row of temporal feature matrix \( \mathbf{X}_t \in \mathbb{R}^{|V| \times d} \). The collection of $\mathbf{X}_t$ across all time forms the overall feature matrix $\mathcal{X}$.

\stitle{Dynamic graph encoder.}
Dynamic graph neural network (DGNN) \cite{wu2020comprehensive} is a mainstream technique for dynamic graph encoding. Given time $t$, for the $l$-th DGNN layer, we aggregate embedding from previous layer to compute the node embedding $\vec{h}^l_{t,v}=$
% {\footnotesize
% \begin{align}\label{eq.DGNN}
%     \mathtt{Aggr}(\mathtt{Fuse}(\vec{h}^{l-1}_{t,v},\mathtt{TE}(t)), \{\mathtt{Fuse}(\vec{h}^{l-1}_{t',u},\mathtt{TE}(t')) : (u,t')\in\bN_v\}),
% \end{align}}
\begin{align}\label{eq.DGNN}
    \mathtt{Aggr}(\mathtt{Fuse}(\vec{h}^{l-1}_{t,v},\mathtt{TE}(t)), \{\mathtt{Fuse}(\vec{h}^{l-1}_{t',u},\mathtt{TE}(t')) : (u,t')\in\bN_v\}),
\end{align}

where $\bN_v$ denotes the set of historical neighbors of $v$, with $(u,t')\in \bN_v$ denoting that $u$ interacted with $v$ at time $t'<t$. $\mathtt{Aggr}(\cdot)$ is an aggregation function. $\mathtt{TE}$ is a time encoder which encodes time interval \cite{cong2023we,rossi2020temporal} as follows:
\begin{align}
    \vec{f}_t=\mathtt{TE}(t) =\frac{1}{\sqrt{d}} [ \cos(\omega_1 t), \sin(\omega_1 t), \ldots, \cos(\omega_{d/2} t), \sin(\omega_{d/2} t) ].
\end{align}
For simplicity, we define the dynamic graph encoder as $\mathtt{DGE}$, the embedding of node $v$ from the final layer as $\vec{h}_{t,v}$.%, which is a row of embedding matrix $\mathbf{H}_t$.

\stitle{Graph prompt learning.}
For dynamic graph prompt learning, previous work \cite{yu2024dygprompt,chen2024prompt} first pre-train a dynamic graph encoder via unsupervised pretext task: $\bL(\Theta)= \mathtt{PRE}(\mathtt{DGE}(G,\mathcal{X}))$, where $\mathtt{PRE}(\cdot)$ denotes pre-training tasks such as link prediction \cite{yu2024dygprompt}, $\Theta$ represents the parameters in $\mathtt{DGE}$, $\bL$ is the pre-training loss. 
The pre-trained model is then adapted to downstream applications via prompt tuning. They design prompts $\mathtt{PRO}$ to modify node and time feature: $\hat{\vec{x}}_{v,t}, \hat{\vec{f}}_t= \mathtt{PRO}(\vec{x}_{v,t},\vec{f}_t)$ and then fed into the pre-trained dynamic graph encoder. For static graph prompt learning methods \cite{yu2024few}, the same paradigm is followed, but without time adaptation.

\stitle{Scope of work.}
In this study, we introduce an event-aware prompt learning framework, \model,  leveraging historical events knowledge for downstream adaptation. We assess the performance of \model\ on two widely studied dynamic graph tasks: temporal link prediction and node classification. Specifically, we evaluate \model\ in data-scarce setting, where only limited labeled data are available for task adaptation. Since in real-world applications, labeled data for node classification is often difficult or costly to obtain \cite{zhou2019meta,yao2020graph}, while link prediction tasks typically involve nodes with sparse interactions \cite{lee2019melu,pan2019warm}.

\begin{figure*}[t]
\centering
\includegraphics[width=0.95\linewidth]{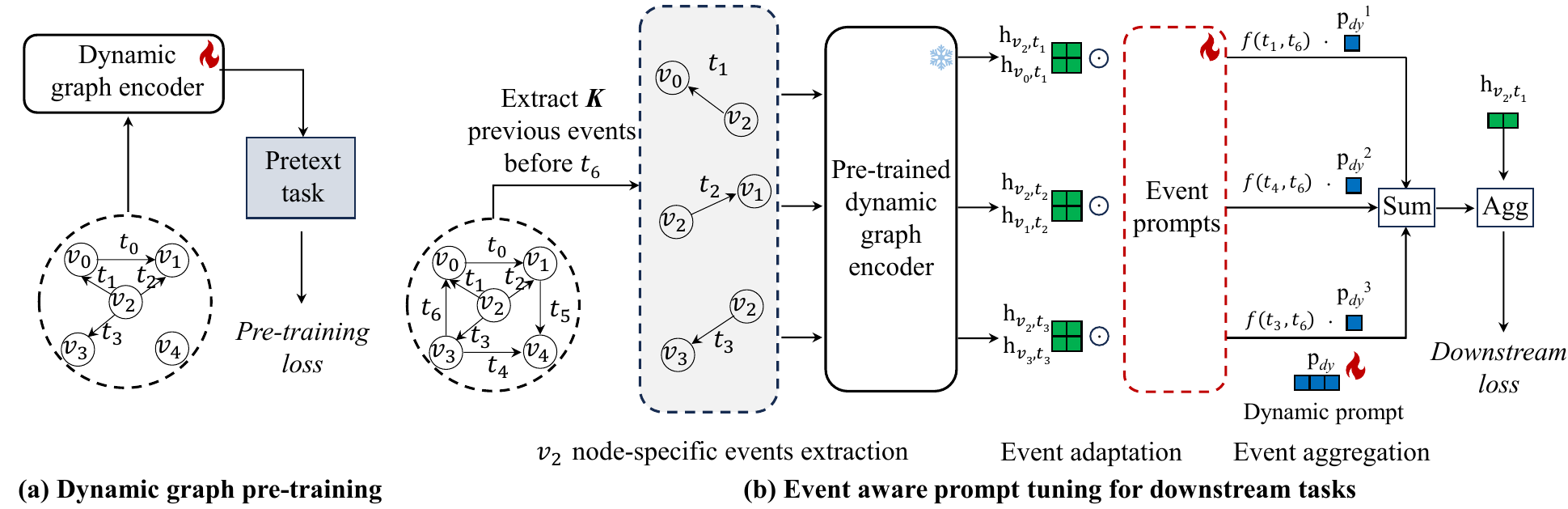}
\caption{Overall framework of \model.}
\Description{Overall framework of the proposed model.}
\label{fig.framework}
\end{figure*}

\section{Proposed Method: \model}\label{sec.model}
In this section, we introduce our proposed model, \model. %We first present an overview of \model, and then detail its core components. 

\subsection{Overall framework}
We illustrate the overall framework of \model\ in Fig.\ref{fig.framework}. Overall, \model\ follows a two-stage pipeline that combines learning a general-purpose dynamic graph encoder and event-aware prompt tuning for efficient downstream adaptation.
First, we pre-train a dynamic graph encoder, as shown in Fig.~\ref{fig.framework}(a). 
The encoder models the dynamic graphs and learns time-dependent node representations via a pretext task, so that the learned representations capture intrinsic temporal and structural patterns and can be transferred to various downstream applications.
Second, given the pre-trained dynamic graph encoder, we propose event-aware prompting to adapt historical event knowledge to downstream tasks through three substages: event extraction, event adaptation, and event aggregation, as shown in Fig.~\ref{fig.framework}(b). 
Specifically, for each node at a target time, we first extract its recent $K$ historical events as a compact event context. 
We then employ lightweight event prompts to modify the resulting event representations, aiming to better align fine-grained event information with the downstream objective while keeping the backbone encoder unchanged. 
Finally, we aggregate the adapted historical events into a history-aware representation via a dynamic prompt together with a time-decay function, which jointly reweights the relative impact of different events based on their learned importance and temporal recency. 
The aggregated representation is then used for downstream prediction and optimized with the task-specific loss.

\subsection{Event extraction} \label{sec.event-extract}
Graph structures evolve over time through a series of temporal events, which record how nodes interact and how local structures change. Such event histories provide valuable signals for downstream prediction, since they reflect both recent dynamics (e.g., short-term intent) and accumulated patterns (e.g., stable preferences) of a node. Therefore, \model\ starts by extracting node-specific historical events as an explicit event context, which serves as the input for subsequent event-aware prompting.

Formally, at time $t$, for a target node $v$, we extract its $K$ most recent historical events before $t$ and denote them as
\begin{align}
\mathcal{E}_{v,t} = \{ E_{v,t}^1, E_{v,t}^2, \dots, E_{v,t}^K \},
\end{align}
where each event represents an interaction involving node $v$ that occurred prior to the query time $t$:
\begin{align}
E_{v,t}^k = (v, u_{v,t}^k, z_{v,t}^k).
\end{align}
Here, $u_{v,t}^k$ is the counterpart node that interacts with $v$ in the $k$-th extracted event, and $z_{v,t}^k$ is the corresponding timestamp. We extract events in reverse chronological order such that
$z_{v,t}^1 \ge z_{v,t}^2 \ge \cdots \ge z_{v,t}^K$ and $z_{v,t}^k < t$, i.e., $E_{v,t}^1$ is the most recent event before time $t$. The event number $K$ is a hyperparameter controlling the history length: a larger $K$ captures richer long-range context but increases computation, while a smaller $K$ focuses on near-term dynamics.
In practice, some nodes may have fewer than $K$ observed events before time $t$, especially for newly appeared or cold-start nodes. In this case, we simply use all available events for node $v$ without padding, and subsequent modules operate on the variable-length event set. This design allows \model\ to handle heterogeneous activity levels across nodes and makes the event extraction stage robust to sparse histories.

\subsection{Event adaptation}\label{sec.event-mod}
After extracting node-specific historical events, the next step is to translate these raw interaction events into representations that are directly useful for a target downstream objective. A key difficulty is that the pre-trained dynamic graph encoder is typically optimized with task-agnostic or link-oriented signals, whereas downstream tasks may require different decision boundaries and emphasize different aspects of interaction evidence. Therefore, rather than fine-tuning the entire encoder, \model\ introduces an \emph{event adaptation} module that performs lightweight, task-aligned modification on top of pre-trained representations. 

\stitle{Event embedding construction.}
For each extracted event $E_{v,t}^k \in \mathcal{E}_{v,t}$, we first compute pre-trained representations for the two endpoint nodes involved in the event. Given the pre-trained dynamic graph encoder (Eq.~\ref{eq.DGNN}), we obtain the time-dependent embeddings for node $v$ and its counterpart $u_{v,t}^k$, denoted as $\vec{h}_v$ and $\vec{h}_{u_{v,t}^k}$, respectively.\footnote{For simplicity, we omit the timestamp in notation; the embeddings are computed under the corresponding temporal context as defined in Eq.~\ref{eq.DGNN}.}
We then fuse the two endpoint embeddings to form an \emph{event embedding}:
\begin{align}\label{eq.event-fuse}
    \vec{e}_{v,t}^k = \mathtt{FUSE}(\vec{h}_v, \vec{h}_{u_{v,t}^k}),
\end{align}
where $\mathtt{FUSE}(\cdot)$ summarizes the interaction evidence of event $E_{v,t}^k$ into a single vector. In general, $\mathtt{FUSE}$ can be implemented with attention \cite{vaswani2017attention}, gated/weighted combinations, or other interaction operators. In this work, we adopt a simple yet stable instantiation by summing the two embeddings, i.e., $\vec{e}_{v,t}^k=\vec{h}_v+\vec{h}_{u_{v,t}^k}$, to focus on the effect of event-aware prompting and to keep the plug-in overhead minimal. As a result, we obtain the event embedding set $\{\vec{e}_{v,t}^1,\vec{e}_{v,t}^2,\dots,\vec{e}_{v,t}^K\}$ for node $v$.

\stitle{Event prompt modification.}
We further adapt each event embedding to better align event-level evidence with the downstream objective:
\begin{align}\label{eq.event-adapt}
    \vec{\hat{e}}_{v,t}^k = \mathtt{EvPro}(\vec{e}_{v,t}^k; \phi),
\end{align}
where $\mathtt{EvPro}$ is the event adaptation mechanism and $\phi$ denotes its learnable parameters. Intuitively, $\mathtt{EvPro}$ acts as an \emph{event prompt} that re-parameterizes the event embedding space for a given task, enabling downstream supervision to selectively amplify or suppress certain embedding dimensions that are most predictive. Importantly, this adaptation is performed \emph{before} event aggregation (Section~\ref{sec.event-agg}), so that the subsequent history representation is constructed from task-aligned event evidence rather than raw pre-trained signals.

In \model, we implement $\mathtt{EvPro}$ with a simple but effective prompt vector:
\begin{align}\label{eq.event-prompt}
    \vec{\hat{e}}_{v,t}^k = \vec{p}_\text{e} \odot \vec{e}_{v,t}^k,
\end{align}
where $\vec{p}_\text{e}$ is a learnable vector with the same dimensionality as $\vec{e}_{v,t}^k$, and $\odot$ denotes element-wise multiplication. This design has two advantages. First, it provides a \emph{dimension-wise gating} mechanism that can directly modulate which latent factors encoded by the backbone are emphasized for the downstream task, while introducing only a negligible number of parameters. Second, since the same prompt is shared across events, it encourages a consistent task-specific re-interpretation of historical evidence, which is crucial when we later aggregate heterogeneous events across time.

\subsection{Event aggregation}\label{sec.event-agg}
Given the event prompt adjusted event embeddings $\{\vec{\hat{e}}_{v,t}^1,\dots,\vec{\hat{e}}_{v,t}^K\}$, we further aggregate them to obtain a compact yet informative historical summary for node $v$ at time $t$. The goal of this stage is to transform a set of heterogeneous historical events into a holistic event knowledge representation for downstream tuning and prediction. Unlike treating all historical events equally, we explicitly account for two complementary factors that commonly arise in dynamic graphs: (i) \textit{temporal recency}, where recent events often carry more immediate relevance, and (ii) \textit{accumulated patterns}, where certain historical events may be disproportionately informative even if they are not the most recent.

\stitle{Recency-aware aggregation.}
A natural prior in dynamic graphs is that events closer to the query time $t$ tend to be more relevant for downstream adaptation at time $t$. To encode this inductive bias, we introduce a time decay function $f(\cdot)$ to weight historical events by their temporal distance:
\begin{align}\label{eq.event-aggregate}
    \vec{\hat{e}}_{v,t} = \sum_{k=1}^K f(z_{v,t}^k, t)\cdot \vec{\hat{e}}_{v,t}^k,
\end{align}
where $z_{v,t}^k$ denotes the timestamp of the $k$-th extracted event. In our implementation, we adopt an exponential decay form:
\begin{align}
    \vec{\hat{e}}_{v,t} = \sum_{k=1}^K \exp(z_{v,t}^k - t)\cdot \vec{\hat{e}}_{v,t}^k,
\end{align}
which smoothly down-weights older events and provides a simple, robust recency prior without introducing additional parameters.

\stitle{Pattern-aware aggregation.}
While recency is broadly useful, it is not always sufficient: some historical events may align more strongly with a node’s current behavior patterns and thus should receive higher weights even if they are not the latest ones. For example, a user may exhibit periodic behaviors (e.g., posting every night), making certain pattern-consistent events more predictive than temporally closer but less relevant ones. To capture such effects in a parameter-efficient way, we introduce a \emph{dynamic prompt} $\vec{p}_{\text{dy}}\in\mathbb{R}^{K}$ that learns an event-importance profile over the extracted history:
\begin{align}
    \vec{\tilde{e}}_{v,t} = \sum_{k=1}^K \vec{p}_{\text{dy}}^k \cdot \vec{\hat{e}}_{v,t}^k,
\end{align}
where $\vec{p}_{\text{dy}}^k$ denotes the weight associated with the $k$-th extracted event. Different from conventional pooling operators (e.g., mean/max) \cite{gholamalinezhad2020pooling}, $\vec{p}_{\text{dy}}$ provides a lightweight mechanism to \emph{learn} which positions in the extracted history tend to be more informative for a downstream objective. This is particularly suitable for our setting because the event set is constructed in a consistent chronological order, allowing the prompt to model non-uniform importance across the event sequence.

% \stitle{Unified historical event knowledge.}
% The above two components are complementary: the time-decay term enforces a recency-aware prior, while the dynamic prompt captures task-driven salience that may deviate from pure recency. Together, they enable \model\ to construct a history-aware representation that reflects both temporally local signals and pattern-consistent historical evidence. In this way, the aggregated embedding $\vec{\tilde{e}}_{v,t}$ summarizes holistic event knowledge for node $v$ and serves as the key historical context for downstream adaptation and prediction.

\subsection{Prompt tuning}\label{sec.model.tuning}
We integrate the historical events embedding with node embedding:
\begin{align}\label{eq.prompt-tuning}
    \vec{\hat{h}}_{v,t}=\vec{{h}}_{v,t}+\vec{\tilde{e}}_{v,t}.
\end{align}

For downstream tuning, we adopt the same loss function as the method into which \model\ is plugged. For example, when integrating with DyGPrompt \cite{yu2024dygprompt}, for temporal link prediction, we define the loss function $\bL(\mathcal{D}; \vec{p}_\text{e}, \vec{p}_{\text{dy}})=$:
\begin{align} \label{eq.pre-train-loss}
     -\sum_{(v, a, b, t) \in \mathcal{D}} \ln \frac{\exp\left(\frac{1}{\tau} \text{sim}(\vec{\hat{h}}_{v,t}, \vec{\hat{h}}_{a,t})\right)}{\exp\left(\frac{1}{\tau} \text{sim}(\vec{\hat{h}}_{v,t}, \vec{\hat{h}}_{b,t})\right)},
\end{align}
where $a$ is a node connected with node $v$ at time $t$, and $b$ is a node disconnected from node $v$ at time $t$. $\tau > 0$ is a temperature hyperparameter.

For temporal node classification, consider a labeled set $\mathcal{D}_\text{down} = \{(v_1, y_1, t_1), (v_2, y_2, t_2), \ldots\}$, where each $v_i$ denotes a node, and $y_i \in Y$ is the class label of $v_i$ at time $t_i$. We define the downstream loss $\bL(\mathcal{D}; \vec{p}_\text{e}, \vec{p}_{\text{dy}})=$
\begin{align}\label{eq.loss}
% \textstyle 
     -\sum_{(v_i, y_i, t_i) \in \mathcal{D}} \ln\frac{\exp\left(\frac{1}{\tau} \text{sim}(\vec{\hat{h}}_{v,t}, \bar{\vec{h}}_{t_i, y_i})\right)}{\sum_{y \in Y} \exp\left(\frac{1}{\tau} \text{sim}(\vec{\hat{h}}_{v,t}, \bar{\vec{h}}_{t_i, y})\right)},
\end{align}
where $\text{sim}(\cdot)$ is a similarity calculation function, here we use cosine similarity. $\bar{{\vec{h}}}_{t_i, y}$ is the class $y$'s prototype embeddings \cite{liu2023graphprompt} at time $t_i$, obtained by averaging the embeddings of examples in class $y$ at time $t_i$.
In all downstream settings, we optimize only the prompt parameters in $\mathtt{EvPro}$ and the event-importance prompt (i.e., $\vec{p}_\text{e}$ and $\vec{p}_{\text{dy}}$), while keeping the pre-trained dynamic graph encoder frozen.

\subsection{Plug-in for dynamic graph learning}
\model\ can integrate with dynamic graph learning methods, including traditional DGNNs, dynamic graph pre-training methods, and graph prompt learning methods.
Specifically, for traditional DGNNs, \model\ directly leverages them as the dynamic graph encoder for pre-training, and then uses the pre-trained DGNN for downstream adaptation. 

For pre-training and prompt learning methods, \model\ follows the same pre-training methods they use and plugs into their downstream adaptation phase. The basic difference between pre-training methods and prompt learning methods is in downstream adaptation phase. For pre-training methods, they generally fine-tune a task head and the pre-trained model. Therefore, we leverage \model\ to integrate historical events knowledge into the pre-trained node embedding, then tune the task head and \model\ for downstream task.

Prompt learning methods generally design prompts to modify node/time feature, as detailed in Sect.~\ref{sec.pre}. Formally, given the prompting method $\mathtt{PRO}(\cdot)$, we first obtain the prompt adjusted-feature $\hat{\vec{x}}_{v,t} = \mathtt{PRO}(\vec{x}_{v,t})$, and then feed it into the dynamic graph encoder to obtain the node embedding ${\vec{h}}_{v,t}$. Next, we use \model\ to extract events and calculate event embeddings (Eq.~\ref{eq.event-fuse}), and then conduct event adaptation (Eq.~\ref{eq.event-adapt}) and event aggregation (Sect.~\ref{eq.event-aggregate}). For prompt tuning, we use the same loss function as the one used by the prompt learning method. The performance of \model\ plugged into DGNNs, dynamic graph pre-training and prompt learning methods is shown in Table~\ref{table.plugin}.

\begin{algorithm}[h]
\small
\caption{\textsc{Event-Aware Prompt Tuning}}
\label{alg.evp}
\begin{algorithmic}[1]
    \Require Pre-trained dynamic graph encoder $\mathtt{DGE}$ with parameters $\Theta_0$; $\mathcal{D},K,\tau$;
    \Ensure Optimized event prompt $\vec{p}_\text{e}$ and dynamic prompt $\vec{p}_{\text{dy}}$
    
    \State $\vec{p}_\text{e}, \vec{p}_{\text{dy}} \leftarrow$ initialization
    
    \While{not converged}
        \For{each training instance in $\mathcal{D}$}
            % \State Parse instance to obtain query node $v$ and query time $t$ (and label/targets required by the downstream task)            
            \State \slash* Event extraction (Section~\ref{sec.event-extract}) *\slash
            \State $\mathcal{E}_{v,t} \leftarrow \{E_{v,t}^1,\dots,E_{v,t}^K\}$ %where $E_{v,t}^k=(v,u_{v,t}^k,z_{v,t}^k)$ and $z_{v,t}^k<t$
            
            % \State \slash* Obtain pre-trained temporal node embeddings (Eq.~\ref{eq.DGNN}) *\slash
            \State $\vec{h}_{v,t} \leftarrow \mathtt{DGE}(v,t;\Theta_0)$ \Comment{Eq.~\ref{eq.DGNN}}
            
            \State \slash* Event adaptation (Section~\ref{sec.event-mod}) *\slash
            \For{each event $E_{v,t}^k \in \mathcal{E}_{v,t}$}
                \State $\vec{h}_{u_{v,t}^k,z_{v,t}^k} \leftarrow \mathtt{DGE}(u_{v,t}^k, z_{v,t}^k;\Theta_0)$
                \State $\vec{e}_{v,t}^k \leftarrow \mathtt{FUSE}(\vec{h}_{v,t}, \vec{h}_{u_{v,t}^k,z_{v,t}^k})$  \Comment{Eq.~\ref{eq.event-fuse}}
                \State $\vec{\hat{e}}_{v,t}^k \leftarrow \vec{p}_\text{e} \odot \vec{e}_{v,t}^k$ \Comment{Eq.~\ref{eq.event-prompt}}
            \EndFor
            
            \State \slash* Event aggregation (Section~\ref{sec.event-agg}) *\slash
            \State $\vec{\tilde{e}}_{v,t} \leftarrow \vec{0}$
            \For{$k=1$ to $|\mathcal{E}_{v,t}|$}
                \State $w_k \leftarrow \exp(z_{v,t}^k-t)\cdot \vec{p}_{\text{dy}}^k$
                \State $\vec{\tilde{e}}_{v,t} \leftarrow \vec{\tilde{e}}_{v,t} + w_k \cdot \vec{\hat{e}}_{v,t}^k$
            \EndFor
            
            \State \slash* Prompt tuning (Section~\ref{sec.model.tuning}) *\slash
            \State $\vec{\hat{h}}_{v,t} \leftarrow \vec{h}_{v,t} + \vec{\tilde{e}}_{v,t}$ \Comment{Eq.~\ref{eq.prompt-tuning}}
            
            \State \slash* Compute downstream loss and update prompts *\slash
            \State Calculate $\bL_{\text{down}}(\mathcal{D};\vec{p}_\text{e},\vec{p}_{\text{dy}})$ \Comment{Eq.~\ref{eq.pre-train-loss},~\ref{eq.loss}} 
            \State Update $\vec{p}_\text{e}, \vec{p}_{\text{dy}}$ by backpropagating $\bL_{\text{down}}(\mathcal{D};\vec{p}_\text{e},\vec{p}_{\text{dy}})$
        \EndFor
    \EndWhile
    
    \State \Return $\vec{p}_\text{e}, \vec{p}_{\text{dy}}$
\end{algorithmic}
\end{algorithm}

\subsection{Algorithm}
% We outline the key steps of \model\ in Algorithm~\ref{alg.evp}. In lines 4--6, for each training instance with query node $v$ at time $t$, we first extract the $K$ most recent historical events $\mathcal{E}{v,t}$ to form a compact event context. In lines 7--12, we compute the pre-trained temporal embeddings using the frozen dynamic graph encoder and construct event embeddings by fusing the endpoint representations for each extracted event; we then perform \emph{event adaptation} by applying the learnable event prompt $\vec{p}e$ to calibrate event embeddings toward the downstream objective. In lines 13--17, we aggregate the adapted events into holistic historical event knowledge: we combine a recency-aware time decay term with the dynamic event-importance prompt $\vec{p}{\text{dy}}$ to reweight and sum historical events, producing $\vec{\tilde{e}}{v,t}$. In line 18, we inject the aggregated event knowledge into the backbone embedding via residual integration to obtain the history-aware representation $\vec{\hat{h}}_{v,t}$ for downstream prediction. Finally, in lines 19--20, we compute the downstream loss using the original objective of the plugged-in method (e.g., temporal link prediction or temporal node classification) and update only the prompt parameters $(\vec{p}e,\vec{p}{\text{dy}})$ by backpropagation, while keeping the pre-trained encoder fixed.

We outline event-aware prompt tuning in Algorithm~\ref{alg.evp}. After initializing $\vec{p}_\text{e}$ and $\vec{p}_{\text{dy}}$ (line 2), we iteratively optimize them on $\mathcal{D}$ (lines 3--20). For each instance $(v,t)$, we extract the $K$ most recent historical events $\mathcal{E}_{v,t}$ (lines 5--6) and compute the embedding $\vec{h}_{v,t}$ using the frozen encoder $\mathtt{DGE}$ (line 7). We then perform \emph{event adaptation} by fusing endpoint embeddings to obtain event embeddings and applying the event prompt $\vec{p}_\text{e}$ to produce $\vec{\hat{e}}_{v,t}^k$ (lines 8--12). Next, we \emph{aggregate} adapted events with a time-decay weight and the dynamic prompt $\vec{p}_{\text{dy}}$ to form $\vec{\tilde{e}}_{v,t}$ (lines 13--17), and inject it into $\vec{h}_{v,t}$ to obtain $\vec{\hat{h}}_{v,t}$ (line 19). Finally, we compute the downstream loss and update only $\vec{p}_\text{e},\vec{p}_{\text{dy}}$ by backpropagation (lines 20--21).

\section{Experiments}
In this section, we conduct experiments to evaluate and analyze the performance of \model.

\subsection{Experimental setup}\label{sec:expt:setup}
\stitle{Datasets.}
We evaluate \model\ on four benchmark datasets. We summarize the datasets in Table~\ref{table.dataset}. 

\begin{itemize}
    \item \textbf{Wikipedia}\footnote{\url{http://snap.stanford.edu/jodie/wikipedia.csv}} captures a month of edits made by contributors to Wikipedia pages \citep{ferschke2011wikipedia}. Building on previous studies \citep{rossi2020temporal,xu2020inductive}, we focus on data from the most frequently edited pages and active contributors, resulting in a temporal graph with 9,227 nodes and 157,474 temporal directed edges. The dynamic labels indicate whether contributors were temporarily banned from editing.
    \item \textbf{Reddit}\footnote{\url{http://snap.stanford.edu/jodie/reddit.csv}} represents an evolving network between posts and users across subreddits, where an edge signifies a user posting content to a subreddit. This dataset contains approximately 11,000 nodes and 700,000 temporal edges, with dynamic labels indicating whether a user has been banned from posting.
    \item \textbf{MOOC}\footnote{\url{http://snap.stanford.edu/jodie/mooc.csv}} consists of student-course interactions on a MOOC platform. In this dataset, nodes represent users and courses, while edges denote user actions on the courses. Dynamic labels indicate whether a student drops out after taking an action.
    \item \textbf{Genre}\footnote{\url{https://object-arbutus.cloud.computecanada.ca/tgb/tgbn-genre.zip}} is a dynamic network connecting users to music genres, where edges represent users listening to specific genres at different times. The dataset includes 1,505 nodes and 17,858,395 temporal edges, with dynamic labels indicating each user's most preferred music genre.
\end{itemize}

% \begin{table}[tbp]
% \centering
% \caption{Summary of datasets.}
% \label{table.dataset}
% \footnotesize
% \resizebox{1\linewidth}{!}{%
% \begin{tabular}{lcccccc}
% \toprule
% \multirow{2}*{Dataset}   & Nodes & Edges & Node & Dynamic & Feature & Time\\ 
%                          & num &num &classes &labels &dimension &span\\\midrule
% Wikipedia & 9,227    & 157,474  & 2 & 217 & 172 & 30 days\\
% Reddit    & 11,000   & 672,447  & 2 & 366 & 172  & 30 days\\
% MOOC      & 7,144    & 411,749  & 2 & 4,066 & 172 & 30 days\\ 
% Genre &1,505 &17,858,395 & 474 & 984 &86 &1,500 days\\
% \bottomrule
% \end{tabular}}
% \end{table}

\begin{table}[tbp]
\centering
\caption{Summary of datasets.}
\label{table.dataset}
\footnotesize
\resizebox{1\linewidth}{!}{%
\begin{tabular}{lccccc}
\toprule
\multirow{2}*{Dataset}   & Nodes & Edges & Node & Feature & Time\\ 
                         & num &num &classes &dimension &span\\\midrule
Wikipedia & 9,227    & 157,474  & 2 & 172 & 30 days\\
Reddit    & 11,000   & 672,447  & 2 & 172  & 30 days\\
MOOC      & 7,144    & 411,749  & 2 & 172 & 30 days\\ 
Genre &1,505 &17,858,395 & 474  &86 &1,500 days\\
\bottomrule
\end{tabular}}
\end{table}

\stitle{Downstream setting.}\label{sec.implementation}
We evaluate the performance of \model\ through temporal link prediction tasks and temporal node classification. Experiments for link prediction are conducted in both transductive and inductive settings. In the transductive setting, nodes in the test set are observed during the pre-training or downstream tuning phase. In contrast, in the inductive setting, nodes in the test set are not observed during pre-training or downstream tuning. 

We follow previous work \cite{yu2024dygprompt} using the same data split and task construction. Specifically, given a series of events ordered by time, the first 80\% of events are used for pre-training. The remaining 20\% of events are set aside for downstream tasks, further divided into 1\%/1\%/18\% subsets. The first 1\% serves as the training pool for downstream prompt tuning, the second 1\% as the validation pool, and the final 18\% for testing. We pre-train a dynamic graph encoder only once for each dataset and use the pre-trained model for all downstream tasks. 

For the four benchmark datasets we used, each edge is sourced from a user node. We randomly sample 30 events from the training pool, ensuring that at least one user from each class is included. 
For link prediction tasks, we treat the sampled user nodes as target instances, and the corresponding destination nodes in the sampled events as positive instances. For instance, given a sampled event $(v, a, t)$, node $a$ serves as the positive instance for the user node $v$. We further sample a destination node $b$ from the training pool as a negative instance, ensuring that $b$ is not connected to user node $v$ at time $t$. For transductive link prediction, we expand the test set by including negative instances. For inductive link prediction, we exclude nodes that have been observed during the pre-training or downstream tuning phases.
For node classification tasks, the labels of the sampled user nodes at the time of the corresponding event are used as the ground truth for downstream prompt tuning. In the test set, all user nodes from the testing events are included.
We repeat the sampling process 100 times to construct 100 distinct tasks for both link prediction and node classification to ensure robust results.

To evaluate the performance of \model, we use the AUC-ROC metric for both link prediction \cite{sun2022self,bei2024cpdg} and node classification \cite{xu2020inductive,rossi2020temporal}. For each task, we run the experiments with five different random seeds. Therefore, for 100 downstream tasks, we obtain 500 results. We report the average and standard deviation of these results in the following part.

\begin{table*}[tbp]
\centering
\caption{AUC-ROC (\%) evaluation of temporal link prediction and node classification.}
\label{table.fewshot}
\addtolength{\tabcolsep}{-1mm}
\resizebox{\textwidth}{!}{%
\begin{tabular}{@{}l|cccc|cccc|cccc@{}}
\toprule
\multirow{2}*{Methods} & \multicolumn{4}{c|}{{Transductive Link Prediction}} & \multicolumn{4}{c|}{{Inductive Link Prediction}} & \multicolumn{4}{c}{{Node Classification}} \\
\multicolumn{1}{c|}{} & {Wikipedia} & {Reddit} & {MOOC} & {Genre} & {Wikipedia} & {Reddit} & {MOOC} & {Genre} & {Wikipedia} & {Reddit} & {MOOC} & {Genre} \\ \midrule\midrule
\method{GCN-ROLAND}       &49.61\text{\scriptsize ±3.12} &50.01\text{\scriptsize ±2.53} &49.82\text{\scriptsize ±1.44} &49.15\text{\scriptsize ±3.74} &49.60\text{\scriptsize ±2.37} &49.90\text{\scriptsize ±1.64} &49.16\text{\scriptsize ±2.48} &47.25\text{\scriptsize ±2.97} &58.86\text{\scriptsize ±10.3} &48.25\text{\scriptsize ±9.57} &49.93\text{\scriptsize ±6.74} &46.33\text{\scriptsize ±3.97} \\
\method{GAT-ROLAND}     &52.34\text{\scriptsize ±1.82} &50.04\text{\scriptsize ±1.98} &55.74\text{\scriptsize ±3.71} &47.69\text{\scriptsize ±2.81} &52.29\text{\scriptsize ±1.97} &49.85\text{\scriptsize ±2.35} &54.01\text{\scriptsize ±2.16} &49.38\text{\scriptsize ±2.72}&62.81\text{\scriptsize ±9.88} &47.95\text{\scriptsize ±8.42} &50.01\text{\scriptsize ±6.34} & 47.26\text{\scriptsize ±3.49}\\%\midrule
\method{TGAT}       &55.78\text{\scriptsize ±2.03} &62.43\text{\scriptsize ±1.86} &51.49\text{\scriptsize ±1.30} &69.11\text{\scriptsize ±3.89} &48.21\text{\scriptsize ±1.55} &57.30\text{\scriptsize ±0.70} &51.42\text{\scriptsize ±4.27} &48.38\text{\scriptsize ±4.72}&67.00\text{\scriptsize ±5.35} &53.64\text{\scriptsize ±5.50} &59.27\text{\scriptsize ±4.43}  &51.26\text{\scriptsize ±2.31}\\
\method{TGN}      &72.48\text{\scriptsize ±0.19} &67.37\text{\scriptsize ±0.07} &54.60\text{\scriptsize ±0.80} &\text{86.46}\text{\scriptsize ±2.84} &74.38\text{\scriptsize ±0.29} &69.81\text{\scriptsize ±0.08} &54.62\text{\scriptsize ±0.72} &87.17\text{\scriptsize ±2.68}&50.61\text{\scriptsize ±13.6} &49.54\text{\scriptsize ±6.23} &50.33\text{\scriptsize ±4.47} & 50.72\text{\scriptsize ±2.31}\\
\method{TREND}      &63.24\text{\scriptsize ±0.71}&80.42\text{\scriptsize ±0.45}&58.70\text{\scriptsize ±0.78} & 52.78\text{\scriptsize ±1.14} 
&50.15\text{\scriptsize ±0.90}&65.13\text{\scriptsize ±0.54}&57.52\text{\scriptsize ±1.01} &45.31\text{\scriptsize ±0.43} &69.92\text{\scriptsize ±9.27} &\text{64.85\scriptsize ±4.71} &66.79\text{\scriptsize ±5.44} & 50.34\text{\scriptsize ±1.62}\\
\method{GraphMixer}  & 59.73\text{\scriptsize ±0.35} & 61.88\text{\scriptsize ±0.11} & 52.42\text{\scriptsize ±1.38}&60.83\text{\scriptsize ±3.25} & 51.34\text{\scriptsize ±0.84} & 57.64\text{\scriptsize ±0.31}&51.16\text{\scriptsize ±2.59} &56.32\text{\scriptsize ±3.08}&  65.43\text{\scriptsize ±4.21} & 60.21\text{\scriptsize ±5.36} & 63.72\text{\scriptsize ±4.98}& 50.15\text{\scriptsize ±1.49}\\
\midrule
\method{DDGCL}       
&54.96\text{\scriptsize ±1.46} &61.68\text{\scriptsize ±0.81} &55.62\text{\scriptsize ±0.32} &68.49\text{\scriptsize ±5.31}
&47.98\text{\scriptsize ±1.11} &55.90\text{\scriptsize ±1.13} &55.18\text{\scriptsize ±2.73} &42.70\text{\scriptsize ±3.26}&65.15\text{\scriptsize ±4.54} &55.21\text{\scriptsize ±6.19} &62.34\text{\scriptsize ±5.13} &50.91\text{\scriptsize ±2.08}\\
\method{CPDG}       
&52.86\text{\scriptsize ±0.64} &59.72\text{\scriptsize ±2.53} &53.82\text{\scriptsize ±1.50} &49.71\text{\scriptsize ±2.64}
&47.37\text{\scriptsize ±2.23} &56.40\text{\scriptsize ±1.17} &53.58\text{\scriptsize ±2.10} &40.01\text{\scriptsize ±3.59}&43.56\text{\scriptsize ±6.41} &65.92\text{\scriptsize ±6.25} &50.32\text{\scriptsize ±5.06} &49.89\text{\scriptsize ±1.34}\\\midrule
\method{GraphPrompt}  
&55.67\text{\scriptsize ±0.26} &67.46\text{\scriptsize ±0.31} &51.07\text{\scriptsize ±0.75} &86.78\text{\scriptsize ±3.14}
&48.46\text{\scriptsize ±0.28} &59.18\text{\scriptsize ±0.49} &50.27\text{\scriptsize ±0.58} &87.45\text{\scriptsize ±2.57}&73.78\text{\scriptsize ±5.62} &60.89\text{\scriptsize ±6.37} &64.60\text{\scriptsize ±5.76} &51.28\text{\scriptsize ±2.43}\\
\method{ProG} & 92.28\text{\scriptsize ±0.21} & 93.32\text{\scriptsize ±0.06}&58.73\text{\scriptsize ±1.58}  &86.24\text{\scriptsize ±2.87} &89.75\text{\scriptsize ±0.28} &90.69\text{\scriptsize ±0.08}&56.42\text{\scriptsize ±1.95}  &85.43\text{\scriptsize ±3.16}&60.86\text{\scriptsize ±7.43} & 68.60\text{\scriptsize ±5.64}&63.18\text{\scriptsize ±4.79} &51.46\text{\scriptsize ±2.38} \\\midrule
\method{TIGPrompt}      
&82.04\text{\scriptsize ±2.03} &83.26\text{\scriptsize ±2.38} &65.00\text{\scriptsize ±4.73} &86.25\text{\scriptsize ±2.43}
&81.75\text{\scriptsize ±1.97} &79.51\text{\scriptsize ±2.58} &64.98\text{\scriptsize ±4.61} &86.19\text{\scriptsize ±3.06}&69.21\text{\scriptsize ±8.88} &67.70\text{\scriptsize ±9.64} &73.90\text{\scriptsize ±6.68} &\underline{51.38}\text{\scriptsize ±2.72}\\
% \method{TGN-TIGPrompt} &44.80\text{\scriptsize ±5.45} &63.75\text{\scriptsize ±5.60} &55.42\text{\scriptsize ±3.60} &50.84\text{\scriptsize ±2.75}
% &82.04\text{\scriptsize ±2.03} &83.26\text{\scriptsize ±2.38} &\underline{65.00}\text{\scriptsize ±4.73} &86.25\text{\scriptsize ±2.43}
% &81.75\text{\scriptsize ±1.97} &79.51\text{\scriptsize ±2.58} &\underline{64.98}\text{\scriptsize ±4.61} &86.19\text{\scriptsize ±3.06}\\\midrule
\method{DyGPrompt}
&\underline{94.33}\text{\scriptsize ±0.12} &\underline{96.82}\text{\scriptsize ±0.06} &\underline{70.17}\text{\scriptsize ±0.75} &\underline{87.02}\text{\scriptsize ±1.63}
&\underline{92.22}\text{\scriptsize ±0.19} &\underline{95.69}\text{\scriptsize ±0.08} &\underline{69.77}\text{\scriptsize ±0.66} &\underline{87.63}\text{\scriptsize ±1.97}
&\underline{82.09}\text{\scriptsize ±6.43} &\underline{74.00}\text{\scriptsize ±3.10} &\underline{77.78}\text{\scriptsize ±5.08} 
&\textbf{52.03}\text{\scriptsize ±2.24}\\
\model
&\textbf{98.47}\text{\scriptsize ±0.80} 
&\textbf{99.85}\text{\scriptsize ±0.14} 
&\textbf{98.16}\text{\scriptsize ±0.54} 
&\textbf{99.90}\text{\scriptsize ±0.02}
&\textbf{98.12}\text{\scriptsize ±0.85} 
&\textbf{99.79}\text{\scriptsize ±0.15} 
&\textbf{97.97}\text{\scriptsize ±0.64} 
&\textbf{99.84}\text{\scriptsize ±0.04}
&\textbf{87.18}\text{\scriptsize ±3.21} 
&\textbf{76.77}\text{\scriptsize ±7.93} 
&\textbf{78.78}\text{\scriptsize ±4.04} 
&\text{50.45}\text{\scriptsize ±0.33}
\\
% TGN-\model &\underline{74.47}\text{\scriptsize ±3.44} &\textbf{74.00}\text{\scriptsize ±3.10} &69.06\text{\scriptsize ±3.89} &\underline{51.97}\text{\scriptsize ±2.16}
% &\textbf{94.33}\text{\scriptsize ±0.12} &\textbf{96.82}\text{\scriptsize ±0.06} &\textbf{70.17}\text{\scriptsize ±0.75} &\textbf{87.02}\text{\scriptsize ±1.63}
% &\textbf{92.22}\text{\scriptsize ±0.19} &\textbf{95.69}\text{\scriptsize ±0.08} &\textbf{69.77}\text{\scriptsize ±0.66} &\textbf{87.63}\text{\scriptsize ±1.97}\
\bottomrule
\hline
\end{tabular}%
}
   \parbox{1\linewidth}{\scriptsize Results are reported in percent. The best method is bolded and the runner-up is underlined.}
\end{table*}

\stitle{Baselines.}
We leverage four state-of-the-art approaches as baselines to assess the effectiveness of \model. 

\noindent (1) \textbf{Traditional DGNNs}
\begin{itemize}
\item \textbf{ROLAND} \citep{you2022roland}: ROLAND extends static GNN architectures to the dynamic graph setting by treating node embeddings across layers as hierarchical states. This design allows it to model the temporal evolution of the graph's structure effectively.
\item \textbf{TGAT} \citep{rossi2020temporal}: TGAT utilizes self-attention mechanisms alongside time encoding inspired by Bochner's theorem from harmonic analysis. It views node embeddings as time-dependent functions, enabling the model to predict embeddings for both unseen and observed nodes as the graph evolves, by stacking TGAT layers.
\item \textbf{TGN} \citep{xu2020inductive}: TGN employs a memory-based approach that updates node representations based on newly arrived events. This method is designed to capture long-term dependencies across time, and it introduces a parallelizable training strategy to improve efficiency.
\item \textbf{TREND} \citep{wen2022trend}: TREND integrates the Hawkes process into GNNs, employing both event-specific dynamics and node-level dynamics to capture the nuanced relationships between individual events and the aggregate influence of events on each node.
\item \textbf{GraphMixer} \citep{cong2023we}: GraphMixer simplifies feature learning by using a basic MLP, where a fixed portion of the parameters is dedicated to encoding temporal information. This approach improves the model's capacity to model temporal dynamics while retaining a lightweight and flexible architecture.
\end{itemize}

\noindent (2) \textbf{Dynamic Graph Pre-training Methods}
\begin{itemize}
\item \textbf{DDGCL} \citep{tian2021self}: DDGCL proposes a self-supervised method for pre-training dynamic graphs by contrasting two temporally adjacent perspectives of the same node identity, enhancing the capture of temporal relationships.
\item \textbf{CPDG} \citep{bei2024cpdg}: CPDG employs a dual contrastive pre-training strategy, integrating both long-term and short-term temporal patterns to create comprehensive node representations that better reflect dynamic graph characteristics.
\end{itemize}

\noindent (3) \textbf{Static Graph Prompting Methods}
\begin{itemize}
\item \textbf{GraphPrompt} \citep{liu2023graphprompt}: GraphPrompt leverages subgraph similarity to seamlessly integrate various pretext and downstream tasks, including link prediction, node classification, and graph classification. It then tunes a learnable prompt tailored to each specific downstream task.
\item \textbf{ProG} \citep{sun2023all}: ProG transforms node- and edge-level tasks into graph-level challenges, proposing the use of prompt graphs that are designed with distinct nodes and structures to effectively guide task-specific learning.
\end{itemize}

\noindent (4) \textbf{Dynamic Graph Prompting Methods}
\begin{itemize}
\item \textbf{TIGPrompt} \citep{chen2024prompt}: TIGPrompt introduces a dynamic prompt generator that produces time-aware prompts for individual nodes, thereby enhancing the adaptability and expressiveness of node embeddings for downstream tasks.
\item \textbf{DyGPrompt} \cite{yu2024dygprompt}: DyGPrompt propose dual prompts and dual condition-nets. It first leverages dual prompts to unify the gap between pre-training and downstream tasks, then conditioned on nodes and time feature to generate conditional prompt, thus adapting the pattern between node and time to downstream tasks.
\end{itemize}

\begin{table*}[tbp] 
    \centering
    \small
    %\caption{AUC-ROC (\%) evaluation of \model\  when used as a plug-in to existing methods.}\vspace{1mm}
    \caption{AUC-ROC (\%) evaluation of \model\  when used as a plug-in to existing methods.}
    \label{table.plugin}%
    \resizebox{1\linewidth}{!}{%
    \begin{tabular}{@{}l|l|ccc|ccc|ccc@{}}
    \toprule
    \multirow{2}{*}{Methods} & {Downstream} &\multicolumn{3}{c|}{Transductive link prediction} &\multicolumn{3}{c|}{Inductive link prediction}&\multicolumn{3}{c}{Node classification} \\
    &Adaptation & {Wikipedia} & {Reddit} & {MOOC} & {Wikipedia} & {Reddit} & {MOOC} & {Wikipedia} & {Reddit} & {MOOC}\\
    \midrule
    \multicolumn{11}{c}{Traditional DGNN}\\
    \midrule
    \multirow{2}{*}{TGAT} 
    & -  
    &55.78\text{\scriptsize ±2.03} &62.43\text{\scriptsize ±1.86} &51.49\text{\scriptsize ±1.30}
    &48.21\text{\scriptsize ±1.55} &57.30\text{\scriptsize ±0.70} &51.42\text{\scriptsize ±4.27}&67.00\text{\scriptsize ±5.35} &53.64\text{\scriptsize ±5.50} &59.27\text{\scriptsize ±4.43}\\
    & \method{+\model}  & \textbf{76.50}\text{\scriptsize ±3.48} & \textbf{92.67}\text{\scriptsize ±1.09} & \textbf{76.24}\text{\scriptsize ±5.90} & \textbf{76.65}\text{\scriptsize ±3.67} & \textbf{91.96}\text{\scriptsize ±0.99} & \textbf{76.59}\text{\scriptsize ±5.68}&\textbf{79.03}\text{\scriptsize ±3.61} &\textbf{67.15}\text{\scriptsize ±4.77} &\textbf{67.41}\text{\scriptsize ±2.75}\\
    \midrule
    \multicolumn{11}{c}{Dynamic graph pre-training}\\
    \midrule
    \multirow{2}{*}{DDGCL} 
    & -   &54.96\text{\scriptsize ±1.46} &61.68\text{\scriptsize ±0.81} &55.62\text{\scriptsize ±0.32} &47.98\text{\scriptsize ±1.11} &55.90\text{\scriptsize ±1.13} &55.18\text{\scriptsize ±2.73}&65.15\text{\scriptsize ±4.54} &55.21\text{\scriptsize ±6.19} &62.34\text{\scriptsize ±5.13}\\
    & \method{+\model} &\textbf{77.05}\text{\scriptsize ±1.79} &\textbf{78.16}\text{\scriptsize ±1.26}&\textbf{64.42}\text{\scriptsize ±2.16}&\textbf{77.12}\text{\scriptsize ±1.78} &\textbf{75.28}\text{\scriptsize ±1.30}&\textbf{64.54}\text{\scriptsize ±2.14}&\textbf{78.50}\text{\scriptsize ±2.69} &\textbf{66.55}\text{\scriptsize ±3.99} & \textbf{68.10}\text{\scriptsize ±3.30} \\
    \midrule
    \multirow{2}{*}{CPDG} 
    & - &52.86\text{\scriptsize ±0.64} &59.72\text{\scriptsize ±2.53} &53.82\text{\scriptsize ±1.50} &47.37\text{\scriptsize ±2.23} &56.40\text{\scriptsize ±1.17} &53.58\text{\scriptsize ±2.10}&43.56\text{\scriptsize ±6.41} &65.92\text{\scriptsize ±6.25} &50.32\text{\scriptsize ±5.06}\\
    & \method{+\model} &\textbf{67.16}\text{\scriptsize ±1.38} &\textbf{67.70}\text{\scriptsize ±2.56}&\textbf{86.93}\text{\scriptsize ±8.52}&\textbf{67.29}\text{\scriptsize ±1.28} &\textbf{70.26}\text{\scriptsize ±2.26}&\textbf{87.50}\text{\scriptsize ±8.43}&\textbf{82.94}\text{\scriptsize ±3.04} &\textbf{67.25}\text{\scriptsize ±4.87} & \textbf{63.75}\text{\scriptsize ±3.65} \\
    \midrule
    \multicolumn{11}{c}{Static graph prompting}\\
    \midrule
    \multirow{2}{*}{GraphPrompt} 
    & -       &55.67\text{\scriptsize ±0.26} &67.46\text{\scriptsize ±0.31} &51.07\text{\scriptsize ±0.75}&48.46\text{\scriptsize ±0.28} &59.18\text{\scriptsize ±0.49} &50.27\text{\scriptsize ±0.58} &73.78\text{\scriptsize ±5.62} &60.89\text{\scriptsize ±6.37} &64.60\text{\scriptsize ±5.76}\\
    & \method{+\model} &\textbf{96.69}\text{\scriptsize ±1.01} &\textbf{84.46}\text{\scriptsize ±3.90}&\textbf{89.65}\text{\scriptsize ±0.80}&\textbf{96.78}\text{\scriptsize ±0.98} &\textbf{92.46}\text{\scriptsize ±2.24}&\textbf{88.48}\text{\scriptsize ±0.89}&\textbf{78.27}\text{\scriptsize ±2.46} &\textbf{66.96}\text{\scriptsize ±3.96} & \textbf{65.49}\text{\scriptsize ±3.32} \\
    \midrule
    \multirow{2}{*}{ProG} 
    & -  & 92.28\text{\scriptsize ±0.21}& \textbf{93.32}\text{\scriptsize ±0.06} & 58.73\text{\scriptsize ±1.58} & 89.75\text{\scriptsize ±0.28} & 90.69\text{\scriptsize ±0.08}&56.42\text{\scriptsize ±1.95}&60.86\text{\scriptsize ±7.43} & 68.60\text{\scriptsize ±5.64}&63.18\text{\scriptsize ±4.79}\\
    & \method{+\model} & \textbf{97.33}\text{\scriptsize ±0.43}&93.07\text{\scriptsize ±1.47}&\textbf{96.02}\text{\scriptsize ±0.49} &\textbf{97.00}\text{\scriptsize ±0.44} &\textbf{95.16}\text{\scriptsize ±0.97} &\textbf{95.19}\text{\scriptsize ±0.56}&\textbf{68.39}\text{\scriptsize ±9.44} & \textbf{71.54}\text{\scriptsize ±4.32}& \textbf{74.27}\text{\scriptsize ±3.51} \\
    \midrule
    \multicolumn{11}{c}{Dynamic graph prompting}\\
    \midrule
    \multirow{2}{*}{TIGPrompt} 
    & -  &82.04\text{\scriptsize ±2.03} &83.26\text{\scriptsize ±2.38} &65.00\text{\scriptsize ±4.73}&81.75\text{\scriptsize ±1.97} &79.51\text{\scriptsize ±2.58} &64.98\text{\scriptsize ±4.61}&69.21\text{\scriptsize ±8.88} &67.70\text{\scriptsize ±9.64}&\textbf{73.90}\text{\scriptsize ±6.68}\\
    & \method{+\model} & \textbf{90.30}\text{\scriptsize ±5.27}& \textbf{94.55}\text{\scriptsize ±1.15}&\textbf{90.42}\text{\scriptsize ±1.08} &\textbf{89.13}\text{\scriptsize ±5.14} &\textbf{93.22}\text{\scriptsize ±1.56} &\textbf{90.38}\text{\scriptsize ±1.00}&\textbf{76.22}\text{\scriptsize ±7.20} & \textbf{70.30}\text{\scriptsize ±4.90}& 72.79\text{\scriptsize ±4.35} \\
    \midrule
    \multirow{2}{*}{DyGPrompt} 
    & -  &94.33\text{\scriptsize ±0.12} &96.82\text{\scriptsize ±0.06} &70.17\text{\scriptsize ±0.75}&92.22\text{\scriptsize ±0.19} &95.69\text{\scriptsize ±0.08} &69.77\text{\scriptsize ±0.66}&82.09\text{\scriptsize ±6.43} &74.00\text{\scriptsize ±3.10} &77.78\text{\scriptsize ±5.08}\\
    & \method{+\model} &\textbf{98.47}\text{\scriptsize ±0.80} &\textbf{99.85}\text{\scriptsize ±0.14} &\textbf{98.16}\text{\scriptsize ±0.54} &\textbf{98.12}\text{\scriptsize ±0.85} &\textbf{99.79}\text{\scriptsize ±0.15} &\textbf{97.97}\text{\scriptsize ±0.64}&\textbf{87.18}\text{\scriptsize ±3.21} 
    &\textbf{76.77}\text{\scriptsize ±7.93} &\textbf{78.78}\text{\scriptsize ±4.04} \\
    \midrule
    \end{tabular}}
     \parbox{1\linewidth}{``-'' refers to continually training, fine-tuning, or prompting on downstream tasks, following their original method without \model. \\
      ``+\model'' refers to integrating \model\ with each approach as introduced in Sect.~\ref{sec.model}.}
\end{table*}

\subsection{Implementation Details}
\stitle{Environment.} The setup used for conducting our experiments is as follows:
\begin{itemize}
    \item Ubuntu 18.04.6 LTS
    \item CPU information: Intel(R) Core(TM) i9-9900X CPU @ 3.50GHz
    \item GPU information: GeForce RTX 4070Ti (12 GB)
\end{itemize}

\stitle{Details of baselines.}
We employ the code made available by the respective authors for open-source baselines. In the case of the non-open-source models, CPDG and TIGPrompt, we develop our own implementations. Each model is carefully tuned based on the configuration parameters suggested in the original papers to ensure peak performance. We use Adam to optimize all methods. The detailed implementation of different baselines are shown as follows:

For the Roland model, both GCN and GAT are configured with a two-layer architecture.

In the case of TGAT and TGN, we sample 20 temporal neighbors per node to update their embeddings.
For TREND, once we sample the neighboring nodes, we apply the Hawkes process to the temporal neighbors, utilizing different time decay factors depending on the timestamp of each event.
For GraphMixer, an MLP is used to process both the input nodes and their positive and negative counterparts. The output is then passed through a series of linear layers for final prediction during training.

For DDGCL, we contrast two temporally adjacent views of each node, using a time-dependent similarity metric and a GAN-style contrastive loss function to evaluate the similarity.
For CPDG, we perform depth-first and breadth-first search strategies to sample neighbors for each node.

For GraphPrompt, we calculate similarity using a 1-hop subgraph.

For TIGPrompt, we utilize a projection-based prompt generator, as this approach has been shown to deliver the best performance according to the literature.

For DyGPrompt, we adopt a dual-layer perceptron with a bottleneck architecture as the condition-net. The hidden dimension of this network is set to 86 for the \textit{Wikipedia}, \textit{Reddit}, and \textit{MOOC} datasets, while it is reduced to 43 for the \textit{Genre} dataset.

For all baselines, the hidden dimension is set to 172 for the \textit{Wikipedia}, \textit{Reddit}, and \textit{MOOC} datasets, whereas it is set to 86 for the \textit{Genre} dataset.

\stitle{Details of \model.}
For our proposed \model, we integrate with DyGPrompt to conduct experiments. We set the number of sampled events as 9 for link prediction tasks, and 3 for node classification tasks.

\subsection{Performance evaluation}
We present the results for \model\ plugged into DyGPrompt \cite{yu2024dygprompt} on temporal link prediction and temporal node classification tasks, and compare its performance with all baselines in Table~\ref{table.fewshot}. We make two major observations:

First, \model\ outperforms all state-of-the-art methods in both temporal link prediction and node classification tasks, underscoring the effectiveness of the event adaptation and event aggregation mechanism in \model. We conduct an ablation study in Table~\ref{table.ablation} to further assess the contribution of core design in \model.   

Second, \model\ demonstrates superior performance compared to current prompt learning methods, which are unable to leverage historical event knowledge for downstream tasks. This further underscores the effectiveness of \model\ in learning event knowledge for downstream adaptation. %We also integrate \model\ into other dynamic graph learning methods and present the results in Table~\ref{table.plugin}, to evaluate the robustness of \model.

\begin{table*}[tb]
    \centering
    \addtolength{\tabcolsep}{-1mm}
    \caption{Ablation study on the effects of key components.}
    \label{table.ablation}%
    \resizebox{0.85\linewidth}{!}{%
    \begin{tabular}{@{}l|ccc|ccc|ccc@{}}
    \toprule
    \multirow{2}*{Methods}
    & \multicolumn{3}{c|}{Transductive Link Prediction} & \multicolumn{3}{c|}{Inductive Link Prediction}&\multicolumn{3}{c}{Node classification} \\
    & Wikipedia & Reddit & MOOC & Wikipedia & Reddit & MOOC & Wikipedia & Reddit & MOOC\\
    \midrule\midrule
    \model\method{-ep} %只有有ep的结果
	&97.72\text{\scriptsize ±1.03} 
	&99.69\text{\scriptsize ±0.23} 
	&93.71\text{\scriptsize ±6.08} 
	&97.30\text{\scriptsize ±1.06} 
	&99.50\text{\scriptsize ±0.32} 
	&92.81\text{\scriptsize ±6.92} 
	&86.91\text{\scriptsize ±6.14} 
	&72.69\text{\scriptsize ±4.16} 
	&75.70\text{\scriptsize ±5.09} \\
	\model\method{-dp} %只有dp的结果
	&97.68\text{\scriptsize ±0.62}  
	&95.53\text{\scriptsize ±1.18} 
	&97.27\text{\scriptsize ±0.27}  
	&97.04\text{\scriptsize ±0.64}  
	&96.39\text{\scriptsize ±0.85}  
	&96.81\text{\scriptsize ±0.31} 
	&84.45\text{\scriptsize ±6.67} 
	&74.41\text{\scriptsize ±5.48} 
	&76.89\text{\scriptsize ±3.93} \\
	\model\method{-td} %指数函数的结果
	&97.31\text{\scriptsize ±0.81} 
	&94.00\text{\scriptsize ±1.22} 
	&95.79\text{\scriptsize ±0.83} 
	&97.24\text{\scriptsize ±0.75} 
	&95.79\text{\scriptsize ±0.83} 
	&93.74\text{\scriptsize ±0.58} 
	&86.35\text{\scriptsize ±6.08} 
	&75.97\text{\scriptsize ±8.00}
	&76.88\text{\scriptsize ±4.34} 
	\\ 
	\method{\model}
	&\textbf{98.47}\text{\scriptsize ±0.80} 
	&\textbf{99.85}\text{\scriptsize ±0.14} 
	&\textbf{98.16}\text{\scriptsize ±0.54} 
	&\textbf{98.12}\text{\scriptsize ±0.85} 
	&\textbf{99.79}\text{\scriptsize ±0.15} 
	&\textbf{97.97}\text{\scriptsize ±0.64} 
	&\textbf{87.18}\text{\scriptsize ±3.21} 
	&\textbf{76.77}\text{\scriptsize ±7.93} 
	&\textbf{78.78}\text{\scriptsize ±4.04} \\
	\bottomrule
	\end{tabular}}
\end{table*}

\subsection{Performance of plug-in}\label{sec.exp.backbones}
\model\ can serve as a plug-in to traditional DGNNs, dynamic graph pre-training methods, and prompt learning methods, aiming to enhance their ability to leverage historical event knowledge for downstream adaptation. The integration of \model\ with these methods is introduced in the last part of model section. Specifically, we integrate \model\ with seven strong-performing baselines, including traditional DGNNs: TGAT \cite{rossi2020temporal}, dynamic graph pre-training methods: DDGCL \cite{tian2021self} and CPDG \cite{bei2024cpdg}, static graph prompt learning methods: GraphPrompt \cite{liu2023graphprompt} and ProG \cite{sun2023all}, and dynamic graph prompt learning methods: TIGPrompt \cite{chen2024prompt} and DyGPrompt \cite{yu2024dygprompt}. We present the original results of these baselines and the results when integrated with \model\ in Table~\ref{table.plugin}. We observe that \model\ consistently improves the performance of these state-of-the-art methods. This demonstrates the effectiveness of \model\ in leveraging events knowledge and its flexibility in being applied to various methods.

\subsection{Ablation studies}\label{sec.exp.ablation}
We compare \model\ with its variants to gain deeper insight into the influence of each component in \model. Specifically, \model-EP refers to the case where, after sampling events, only the event prompt is employed for event adaptation, and events embeddings are directly summed without event aggregation mechanism. \model-DP denotes the scenario where event adaptation is not conducted, and the dynamic prompt is directly trained for event aggregation without time decay function. In contrast, \model-TD represents the use of time decay function without dynamic prompt for event aggregation.
We present the results of these variants in Table~\ref{table.ablation} on \textit{Wikipedia}, \textit{Reddit} and \textit{MOOC}, and make the following observations.

First, \model\ consistently outperforms its variants, demonstrating that the event prompt in the events adaptation phase, along with the dynamic prompt and time decay function in the event aggregation phase, are essential for effectively leveraging historical events knowledge for downstream tasks. This further highlights the necessity of learning events knowledge.

Second, simply using time decay function for event aggregation is not sufficient, as \model-TD generally performs worse than \model-EP and \model-DP. While time constraints weight the importance of different events based on the intuition that more recent events should be more important for the current time, they alone do not fully capture the complexity of historical events knowledge, since some previous events may exhibit patterns similar to the user’s current behavior. This emphasizes the necessity of applying a dynamic prompt for adaptively event aggregation.

Third, event adaptation proves to be beneficial. As observed, in both transductive and inductive link prediction tasks, \model-EP outperforms \model-DP on \textit{Wikipedia} and \textit{Reddit}. \model-EP also shows an advantage in the node classification task on \textit{Wikipedia}. This demonstrates that event adaptation can effectively capture fine-grained events characteristics and modify them to better adapt to downstream tasks.

% \begin{figure}[t]
% \centering
% \includegraphics[width=1\linewidth]{figures/visilization-embedding.pdf}
% \caption{Visualization of output embedding space of nodes.}
% \label{fig.visualization}
% \end{figure}

% \begin{figure}[t]
% \centering
% \includegraphics[width=1\linewidth]{figures/para.pdf}
% \caption{Sensitivity of $\alpha$.}
% \label{fig.para}
% \end{figure}

% \begin{figure*}[t]
% \centering
% \begin{minipage}[b]{0.7\textwidth}
% \centering
% \includegraphics[width=0.75\linewidth]{figures/visilization-embedding.pdf}
% \vspace{-3mm}
% \caption{Visualization of output embedding space of nodes.}
% \label{fig.visualization}
% \end{minipage}%
% \begin{minipage}[b]{0.3\textwidth}
% \centering
% \includegraphics[width=0.8\linewidth]{figures/para.pdf}
% \vspace{-6mm}
% \caption{Sensitivity of $\alpha$.}
% \label{fig.para}
% \end{minipage}
% \end{figure*}

\subsection{Hyperparameter sensitivity}\label{exp.para}
\begin{figure}[t]
\centering
\includegraphics[width=0.7\linewidth]{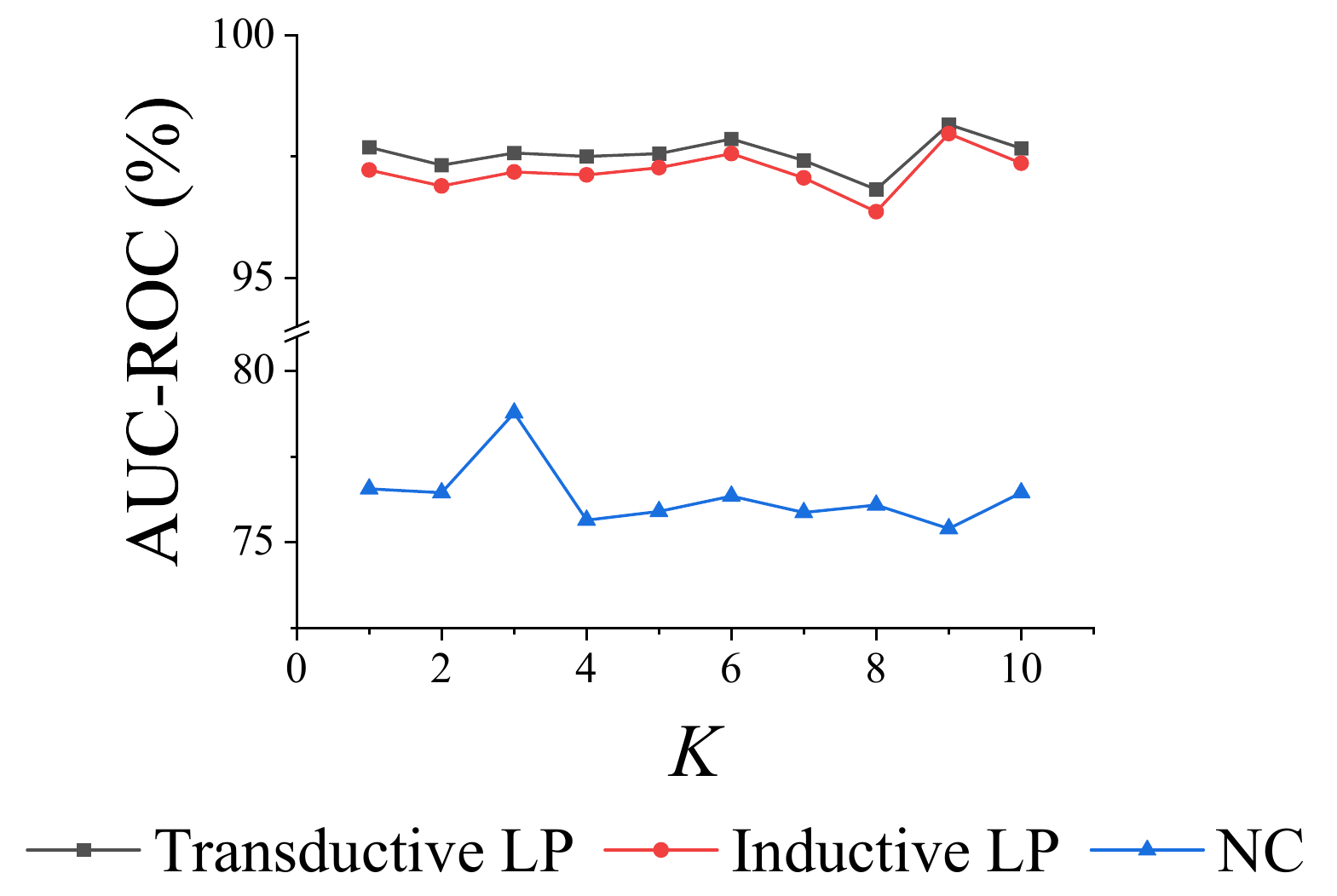}
\caption{Sensitivity of $K$.}
\Description{evaluate the impact of the number of extracted events
$K$ on MOOC, presenting the performance on transductive link pre-
diction (LP), inductive link prediction, and node classification (NC)
tasks}
\label{fig.para}
\end{figure}

We further evaluate the impact of the number of extracted events $K$ on \textit{MOOC}, presenting the performance on transductive link prediction (LP), inductive link prediction, and node classification (NC) tasks in Fig.~\ref{fig.para}. We observe that for both transductive and inductive link prediction, the performance follows a similar pattern: it generally increases as more events are extracted, but then decreases as $K$ continues to increase.
For node classification, the performance initially increases as $K$ increases, reaching a peak at $K=3$. Beyond this point, with more extracted events, the performance shows little change. This may suggest that three historical events are sufficient to capture the necessary historical knowledge for downstream adaptation.
Therefore, in our experiments, we set $K=9$ for link prediction and $K=3$ for node classification.

\section{Conclusions}
In this paper, we propose an event-aware prompt learning method for dynamic graphs, which can serve as a plug-in to enhance prompt learning methods' ability to leverage historical event knowledge. The proposed method, \model, first extracts historical events for each node. We then introduce an event adaptation mechanism to capture the fine-grained characteristics of these events for downstream adaptation. Additionally, we propose an event aggregation mechanism to fuse historical events knowledge to enhance node embedding. Finally, we conduct extensive experiments on four public datasets, demonstrating the effectiveness of \model\ in leveraging historical events knowledge and its robustness as a plug-in to present dynamic graph learning methods.

%\begin{acks}
%To Robert, for the bagels and explaining CMYK and color spaces.
%\end{acks}

\clearpage
\newpage
\bibliographystyle{ACM-Reference-Format}
\bibliography{references}

% \input{sec-appendix}

%\appendix
%\input{sec-appendix}

\end{document}